\documentclass[10pt,twocolumn,letterpaper]{article}

\usepackage{wacv}              

\usepackage{scalerel,xparse}
\usepackage{graphicx}
\usepackage{amsmath}
\usepackage{amssymb}
\usepackage{booktabs}
\usepackage{amsfonts}
\usepackage{booktabs}       

\DeclareMathOperator*{\argmin}{arg\,min}
\usepackage{soul}

\usepackage{algorithm}
\usepackage{algorithmic}
\usepackage{subcaption}
\usepackage{listings}
\usepackage{diagbox}
\usepackage{colortbl}
\usepackage{nicefrac}       
\usepackage{microtype}      
\usepackage{xcolor} 

\usepackage[accsupp]{axessibility}

\newcommand{\cut}[1]{}
\definecolor{codegreen}{rgb}{0,0.6,0}
\definecolor{codegray}{rgb}{0.5,0.5,0.5}
\definecolor{codepurple}{rgb}{0.58,0,0.82}
\definecolor{backcolour}{rgb}{0.95,0.95,0.92}

\lstdefinestyle{mystyle}{
    backgroundcolor=\color{backcolour},   
    commentstyle=\color{codegreen},
    keywordstyle=\color{magenta},
    numberstyle=\tiny\color{codegray},
    stringstyle=\color{codepurple},
    basicstyle=\ttfamily\footnotesize,
    breakatwhitespace=false,         
    breaklines=true,                 
    captionpos=b,                    
    keepspaces=true,                 
    numbers=left,                    
    numbersep=5pt,                  
    showspaces=false,                
    showstringspaces=false,
    showtabs=false,                  
    tabsize=2
}

\usepackage[pagebackref,breaklinks,colorlinks]{hyperref}
\hypersetup{
  colorlinks   = true, 
  urlcolor     = blue, 
  linkcolor    = red, 
  citecolor   = blue 
}

\usepackage[capitalize]{cleveref}
\crefname{section}{Sec.}{Secs.}
\Crefname{section}{Section}{Sections}
\Crefname{table}{Table}{Tables}
\crefname{table}{Tab.}{Tabs.}

\def\wacvPaperID{2077} 
\def\confName{WACV}
\def\confYear{2025}

\begin{document}

\title{\textit{MemControl}: Mitigating Memorization in Diffusion Models via Automated Parameter Selection}


\author{Raman Dutt$^{1}$, Ondrej Bohdal$^{1}$, Pedro Sanchez$^{1}$, Sotirios A. Tsaftaris$^{1}$, Timothy Hospedales$^{1,2}$ \\
$^{1}$The University of Edinburgh \ $^{2}$Samsung AI Center, Cambridge \\
\texttt{\{raman.dutt,ondrej.bohdal,pedro.sanchez,s.tsaftaris,t.hospedales\}@ed.ac.uk} }

\maketitle


\begin{figure*}[ht]
  \centering
  \includegraphics[width=0.6\linewidth]{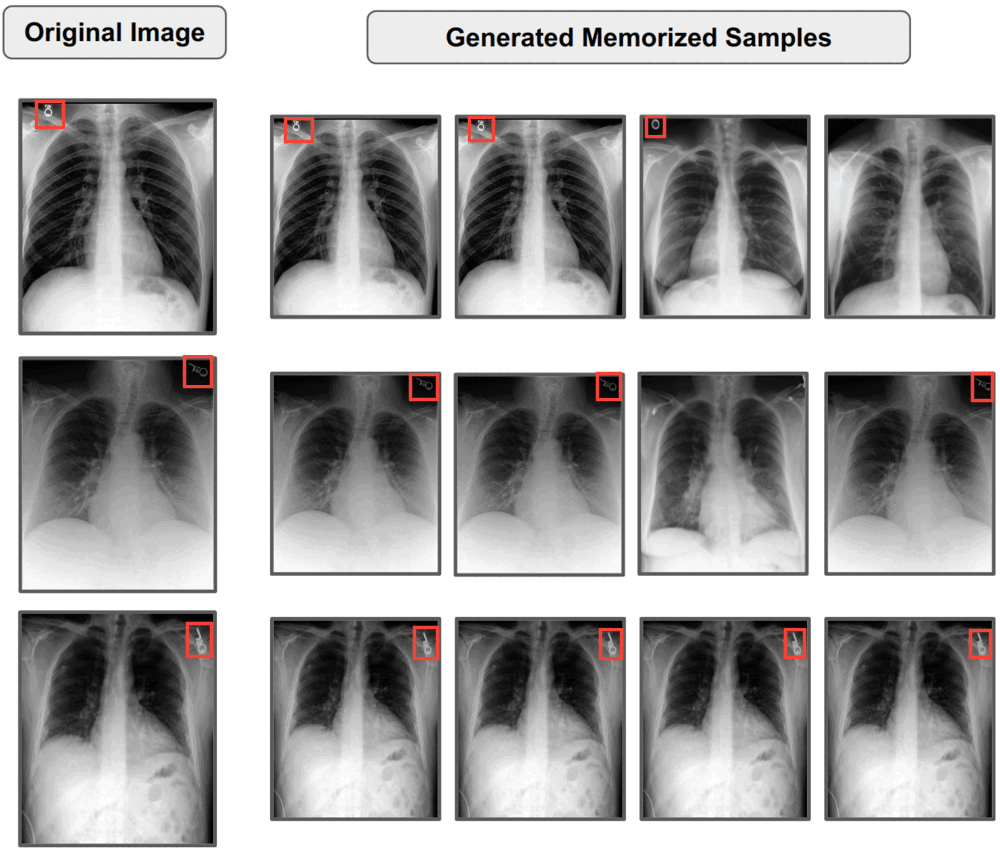}
  \caption{Conventional full fine-tuning results in the generation of nearly identical images across different seeds for the same text prompt. Memorization is evident through the replication of artefacts (red squares), which notably occurs with high precision and is consistently observed in the generated images. This replication can lead to patient information leakage and potential re-identification. Column 1 displays the original training images, while columns 2-5 show the closest generated samples across different seeds.
}
  \label{fig:memorized_samples}
\end{figure*}

\begin{abstract}   
   Diffusion models excel in generating images that closely resemble their training data but are also susceptible to data memorization, raising privacy, ethical, and legal concerns, particularly in sensitive domains such as medical imaging. We hypothesize that this memorization stems from the overparameterization of deep models and propose that regularizing model capacity during fine-tuning can mitigate this issue. Firstly, we empirically show that regulating the model capacity via Parameter-efficient fine-tuning (PEFT) mitigates memorization to some extent, however, it further requires the identification of the exact parameter subsets to be fine-tuned for high-quality generation. To identify these subsets, we introduce a bi-level optimization framework, \textit{MemControl}, that automates parameter selection using memorization and generation quality metrics as rewards during fine-tuning. The parameter subsets discovered through MemControl achieve a superior tradeoff between generation quality and memorization. For the task of medical image generation, our approach outperforms existing state-of-the-art memorization mitigation strategies by fine-tuning as few as 0.019\% of model parameters. Moreover, we demonstrate that the discovered parameter subsets are transferable to non-medical domains. Our framework is scalable to large datasets, agnostic to reward functions, and can be integrated with existing approaches for further memorization mitigation. To the best of our knowledge, this is the first study to empirically evaluate memorization in medical images and propose a targeted yet universal mitigation strategy. The code is available at \href{https://github.com/Raman1121/Diffusion_Memorization_HPO}{this https URL}.

\end{abstract}


\section{Introduction}

Diffusion models \cite{sohl2015deep, song2021denoising, ho2020denoising, song2021scorebased} have proven highly effective in generating high-quality data across various modalities, including images \cite{dhariwal2021diffusion, karras2019style, ramesh2022hierarchical, rombach2022high}, audio \cite{kim2022guided,huang2023make}, and graphs \cite{xu2022geodiff, vignac2023digress}. They power popular products like \textit{Midjourney}, \textit{Stable Diffusion}, and \textit{DALL-E}, enabling image generation based on user prompts. The appeal of these models lies in their ability to create high-quality novel images that capture real-world complexities, with applications ranging from data augmentation for more accurate recognition systems \cite{trabucco2024effective} to academic research on disease progression \cite{kumar2024monitoring}. A key use case is sharing synthetic data to circumvent privacy restrictions, driving progress while addressing privacy concerns. However, this utility is compromised if diffusion models inadvertently disclose training data by generating near-replicas—a phenomenon known as \textit{memorization}, which raises significant legal and ethical concerns \cite{butterick_2023} and identifies diffusion models as potentially the least private form of generative models \cite{carlini2023extracting}.

\textbf{Significance in Medical Image Analysis: \quad }Privacy risks related to memorization are especially significant in the context of healthcare applications.
The risks include exact replication of images, potentially disclosing information about the patient \cite{yoon2020anonymization} and making it possible to conduct re-identification attacks \cite{fernandez2023privacy}. In re-identification attacks, models that can identify whether two images belong to the same patient can allow an attacker with partial information to link a memorized synthetic image to the associated patient. \textcolor{black}{Despite of the significant threat that memorization poses in the medical domain, very limited number of studies have advocated for this issue \cite{fernandez2023privacy,jegorova2022survey,akbar2023beware,dar2023investigating}. Furthermore, to the best of our knowledge, none of the previous works have investigated the efficacy of existing training time mitigation strategies. To bridge this gap, we focus our work on the task of medical image generation and introduce a new training-time memorization intervention method.} 

The growing concern for training data memorization has inspired the development of methods to mitigate memorization. These methods include training-time interventions such as de-duplication \cite{webster2023duplication,somepalli2023understanding} and noise regularizers \cite{somepalli2023diffusion}, as well as inference-time mitigations such as token-rewriting \cite{somepalli2023understanding,wen2024detecting}. \textcolor{black}{Firstly, de-duplication is not possible in the specific case of medical images due to high-similarity between samples. Furthermore, a key limitation of the fast and effective token-rewriting methods is their vulnerability to white-box attacks. An attacker with access to the model can simply disable the token-rewriting during inference to extract private data.}


In this paper, we introduce a novel training-time intervention to address memorization in diffusion models. We hypothesize that the overcapacity of large neural networks contributes to memorization and explore whether regularizing model capacity can mitigate this issue. \textcolor{black}{While similar hypotheses have been investigated in previous work, our approach differs in terms of the task (text-to-image generation as opposed to language generation \cite{hong2024decoding, stoehr2024localizing}, image classification \cite{maini2023can}), the mechanism for regulating capacity (PEFT as opposed to model compression \cite{hong2024decoding}, localization \cite{maini2023can, stoehr2024localizing, chavhan2024memorized}), and the fine-tuning paradigm employed \cite{ghalebikesabi2023differentially}.}

In domain-specific diffusion models, such as those used for medical imaging, the typical workflow involves starting with a general-purpose diffusion model and fine-tuning it with task-specific data \cite{chambon2022roentgen}. We posit that there is a trade-off between the extent of fine-tuning and the potential for the model to memorize in-domain medical images. If no fine-tuning is performed, the pre-trained model may not replicate specific in-domain data but will also struggle to generate high-quality domain-specific images. Conversely, extensive fine-tuning will push the model to generate high-quality in-domain images but also increase memorization risk. We investigate whether it is possible to strike an effective balance, constraining the model enough to produce high-quality generations while minimizing memorization.

To this end, we leverage parameter-efficient fine-tuning (PEFT) \cite{hu2022lora, kopiczko2024vera, liu2024dora, lian2022scaling, han2023svdiff, xie2023difffit}, a set of methods that selectively update specific subsets of parameters during the fine-tuning process. Each PEFT method's unique approach to freezing versus updating parameters results in a different balance between generation quality and memorization. Our study aims to identify which PEFT strategies can best manage this trade-off, ensuring effective specialization of pre-trained models with minimal risk of memorization. We introduce a bi-level optimization framework designed to search for the optimal set of parameters to update, striking a balance between high-quality generation and minimal memorization. Our empirical results on the MIMIC medical imaging dataset \cite{johnson2016mimic} demonstrate a significantly improved quality-memorization tradeoff compared to recent state-of-the-art memorization mitigation techniques \cite{somepalli2023understanding, wen2024detecting}. While our method is fast and robust during inference, it does incur additional computational cost during training. We show the potential to mitigate this cost by transferring high-quality updatable parameter subsets across different learning problems. This suggests our discovered fine-tuning strategy could be applied off-the-shelf in new applications, thus providing a versatile and effective solution for generative tasks in sensitive domains.

The key contributions of our work are summarised as follows: \textcolor{black}{\fbox{\textbf{(1)}}} \textcolor{black}{We conduct the first study that empirically evaluates memorization mitigation mechanisms for medical images, establishing a fair benchmark for future studies;} \textcolor{black}{\fbox{\textbf{(2)}}} Empirically demonstrating that reducing model capacity significantly mitigates memorization; \textcolor{black}{\fbox{\textbf{(3)}}} Highlighting the delicate balance between memorization and generalization and how it is influenced by model capacity (Sec. \ref{sec:capacity_mediates}); \textcolor{black}{\fbox{\textbf{(4)}}} Proposing a bi-level optimization framework that automates parameter (capacity) selection to optimize both generative quality and memorization mitigation (Sec. \ref{sec:HPO_optimization}); and \textcolor{black}{\fbox{\textbf{(5)}}} Showcasing that the parameter subsets discovered are transferable across different datasets and domains (Sec. \ref{sec:transferability}).



\section{Related Work}

\subsection{Memorization in Diffusion Models}  
Memorization in deep generative models has been explored in several contexts, including training data extraction \cite{carlini2021extracting,carlini2023extracting}, content replication \cite{somepalli2023diffusion} and data copying \cite{somepalli2023understanding}.
Memorization also occurs in the medical domain and has been observed for brain MRI and chest X-ray data \cite{akbar2023beware, dar2023investigating}, posing risks for example via patient re-identification \cite{fernandez2023privacy}.
Given the significant risks associated with memorization, several strategies have been proposed to mitigate it. Based on evidence that duplicate samples increase memorization, the most straightforward approach involves de-duplicating the training set \cite{carlini2023extracting,webster2023duplication,naseh2023memory}. However, such approach has been  shown to be insufficient  \cite{somepalli2023understanding}. In text-to-image generative models, text conditioning significantly influences memorization and several mitigation strategies have been proposed based on this observation \cite{somepalli2023understanding,wen2024detecting,ren2024unveiling}.

Memorization mitigation strategies can be categorized into training-time and inference-time methods. Training-time methods include augmenting text input as shown in \cite{somepalli2023understanding}, and can be implemented for example via (1) using multiple captions per image, (2) adding Gaussian noise to text embeddings, (3) adding random words or numbers to text prompts. Among these, adding random words (\textit{RWA}) is shown to be the most effective. Another method, \textit{threshold mitigation}, is based on differences in model output magnitudes between memorized and non-memorized prompts, enabling fast detection and mitigation during training \cite{wen2024detecting}. \cite{ren2024unveiling} have introduced a strategy examining cross-attention score differences, observing higher attention concentration on specific tokens in memorized prompts. Other studies have explored injecting a deliberate memorization signal via a watermark and detecting it in generated images with a trained classifier \cite{wang2023diagnosis} to prevent unintended data usage during training \cite{Radioactive_data}.

\subsection{Parameter-Efficient Fine-Tuning}

Models trained on vast online datasets \cite{schuhmann2022laion} typically lack domain-specific synthesis capabilities, for example they may not be able to generate medical chest X-ray data \cite{chambon2022roentgen,dutt2022automatic}. Consequently, additional fine-tuning following the \textit{pre-train fine-tune} paradigm is necessary. While conventional methods update all parameters via full fine-tuning (FT) \cite{chambon2022roentgen}, recent advancements introduce parameter-efficient alternatives \cite{kopiczko2024vera,liu2024dora,han2023svdiff,xie2023difffit}, achieving comparable performance without full model updates. 

PEFT strategies freeze most pre-trained model parameters and fine-tune either existing parameters (selective PEFT) or introduce and fine-tune new ones (additive PEFT). Selective PEFT methods include attention tuning \cite{touvron2022three}, bias tuning \cite{bitfit}, and normalization tuning \cite{basu2023strong}, while additive PEFT methods include LoRA \cite{hu2021lora}, SSF \cite{lian2022scaling}, SV-DIFF \cite{han2023svdiff}, DiffFit \cite{xie2023difffit} and AdaptFormer \cite{chen2022adaptformer} among others, catering to various visual tasks. PEFT matches or exceeds conventional full fine-tuning  \cite{zhai2019large}, including on medical imaging  \cite{dutt2023parameter}. For text-to-image generation, strategies such as SV-DIFF \cite{han2023svdiff} and DiffFit \cite{xie2023difffit} have been proposed, and they excel in both in-domain and out-of-domain settings \cite{dutt2023parameter}. In addition to performance evaluation, bi-level optimization frameworks based on PEFT have been shown to enhance fairness and reduce bias in discriminative tasks \cite{dutt2024fairtune}.

\section{Methodology}

\subsection{Preliminaries}
\label{sec:metrics}
Diffusion models use a forward and reverse diffusion process. A forward diffusion process involves a fixed Markov chain spanning $T$ steps, where each step introduces a pre-determined amount of Gaussian noise. The noise is iteratively added to a data point \(x_0\) drawn from the real data distribution \(q(x_0)\) as described by:
\begin{align} \label{eqn:forward}
    q\left(x_t \mid x_{t-1}\right)=\mathcal{N}(x_t ; \sqrt{1-\beta_t} x_{t-1}, \beta_t \mathbf{I}),
\end{align}
where \(\beta_t\) represents the scheduled variance at step \(t\). Closed-form expression for this process is:
\begin{equation} \label{eqn:step2}
    x_t = \sqrt{\overline{\alpha}_t}x_0 + \sqrt{1-\overline{\alpha_t}}\epsilon,
\end{equation}
where $\overline{\alpha}_{t}=\prod_{i=1}^{t}(1-\beta_{t})$.

In the reverse diffusion process, a Gaussian vector \(x_T\) sampled from \(\mathcal{N}(0, 1)\) undergoes denoising so that it can be mapped onto an image \(x_0\) that belongs to the distribution \(q(x)\). At each denoising step, a trained noise predictor \(\epsilon_\theta\) removes the noise \(\epsilon_\theta(x_t)\) added to \(x_0\). The formulation for estimating \(x_{t-1}\) is:
\begin{equation}
    x_{t-1} = \sqrt{\overline{\alpha}_{t-1}}\hat{x}_0^t + \sqrt{1 - \overline{\alpha}_{t-1}}\epsilon_{\theta}(x_t),
\end{equation}
where
\begin{equation}
    \hat{x}_0^t = \frac{x_t - \sqrt{1 - \overline{\alpha}_{t-1}}\epsilon_{\theta}(x_t)}{\sqrt{\overline{\alpha}_t}}.
\end{equation}

\subsection{Memorization and Image-Quality Metrics} \label{sec:metrics}
Our goal is to learn the optimal parameter subset for fine-tuning, given a PEFT method to limit the amount of memorization while maintaining generated image quality. There are various metrics used in the community to measure both memorization \cite{carlini2023extracting,somepalli2023diffusion} and generation quality \cite{bannur2023learning}, and we introduce the metrics that we use in this paper next.

\textbf{Memorization: Nearest Neighbour Distance}\quad The simplest definition of memorization considers an example $\hat{x}$ drawn from the model to be memorized if there is a training example $x$ such that $l(x,\hat{x})<\delta$ for some distance metric $l$ and threshold $\delta$. This means a training example has been memorized if the model produces a near-replica as a sample. A natural metric for the overall degree of memorization of a model is the average nearest neighbour distance between each synthetic sample and its closest example in the training set. For $N$ samples $\{\hat{x}_i\}$ drawn from the model and a training set $\{x_j\}$, we follow \cite{bai2021training} and quantify memorization as the average minimum cosine distance (AMD): 

\begin{equation}
d^{amd} = \frac{1}{N} \sum_{i=1}^{N} \min_{j} l_{\cos}(x_i,x_{j}).    
\end{equation}

\textbf{Memorization: Extraction Attack}\quad We also evaluate memorization using an image extraction attack described in \cite{carlini2023extracting}. The attack involves two main steps: (1) Generating numerous image samples for a given prompt, and (2) Conducting membership inference to identify memorized images. In \cite{carlini2023extracting}, membership inference is performed using a heuristic that constructs a graph of similar samples and identifies sufficiently large cliques—sets of samples that are all similar to each other. Samples from these cliques are likely to be memorized examples. Similarity is measured using a modified $\ell_2$ distance, defined as the maximum $\ell_2$ distance across corresponding tiles of the two compared images. A smaller \textit{number of extracted images} translates to less memorization, and we measure it during evaluation.

\textbf{Memorization: Denoising Strength Heuristic}\quad A limitation of the two measures above is that they are comparatively inefficient to compute. An efficient proxy for memorization based on the magnitude of the diffusion noise prediction was introduced in \cite{wen2024detecting}. It represents the discrepancy between model predictions conditioned on a prompt embedding $e_{p}$ and an empty string embedding $e_{\phi}$ averaged across multiple time steps. The detection metric can be formulated as follows: 

\begin{equation}
    d^{mem} = \frac{1}{T}\sum_{t=1}^T ||\epsilon_{\theta}(x_t, e_p) - \epsilon_{\theta}(x_t, e_{\phi})||_2\label{eq:searchMetric},
\end{equation}

for a noise-predictor model $\epsilon_{\theta}$ trained on input samples with Gaussian noise $x_t$ at time-step $t$, incorporating prompt embedding $e_p$, empty string embedding $e_{\phi}$, and $T$ time steps. A higher value of $d^{mem}$ indicates more substantial memorization \cite{wen2024detecting}.

\textbf{Quality: Fréchet Inception Distance (FID)}\quad  FID score assesses the image generation quality by computing a multivariate normal distribution between real and synthetic images estimated from the features of a pre-trained image encoder. 
We use a DenseNet-121 \cite{huang2017densely} model that is pre-trained on chest X-rays \cite{torchxrayvision} to compute the features.

\textbf{Quality: BioViL-T Score:} BioViL-T \cite{bannur2023learning} is a state-of-the-art vision-language model specifically trained for analyzing chest X-Rays and radiology reports. The model returns a joint score denoting the clinical correlation between an image and a text caption. A greater similarity score indicates a higher quality of generated images with better textual guidance.

\subsection{Parameter-Efficient Fine-Tuning}
The key idea of PEFT is to fine-tune a very small subset of model parameters $\phi \subset \theta$, i.e. $|\phi| \ll |\theta|$. In our framework, the selection of which parameters are fine-tuned is represented using a binary mask $\omega$ that is applied on the model parameters $\theta$. Each element of $\omega$ determines if a specific parameter at a specific location in the model should be frozen or fine-tuned. The parameters of the pre-trained model are initially $\theta_0$, and after fine-tuning on data $\mathcal{D}^{train}$ they change by $\Delta\phi$ compared to their initial values. The fine-tuning process can then be defined via
$$\Delta\phi^* = \argmin_{\Delta\phi} \mathcal{L}^{base} \left( \mathcal{D}^{train}; \theta_0 + \omega \odot \Delta\phi  \right),$$
where $\mathcal{L}^{base}$ is the mean squared error in the context of image generation problems.

We aim to determine the optimal mask $\omega$ that results in fine-tuning with minimal memorization and the highest image generation quality.

\subsection{Our Method: Optimizing PEFT for Mitigating Memorization}

Our method employs a multi-objective bi-level optimization formulation. It trains the model in the inner loop assuming a particular PEFT mask that specifies which diffusion model parameters to freeze or learn. The PEFT mask is optimized in the outer loop so as to obtain a diffusion model with low memorisation and high generation quality. This process is formalized using Eq.~\ref{eq:bilevel-obj} and Algorithm~\ref{alg:hpo}:

\begin{equation}
\label{eq:bilevel-obj}
\begin{aligned}
\omega^* &= \underset{\omega}{\operatorname{argmin}} \, 
\mathcal{L}^{outer}\left(d^{mem}\left(\mathcal{D}^{mem}; \Delta\phi^*\right), 
d^{fid}\left(\mathcal{D}^{val}; \Delta\phi^*\right)\right) \\
&\text{such that} \quad 
\Delta\phi^* = \argmin_{\Delta\phi} \mathcal{L}^{base} \left( 
\mathcal{D}^{train}; \theta_0 + \omega \odot \Delta\phi  \right),
\end{aligned}
\end{equation}

where $\mathcal{L}^{base}$ is the usual diffusion model loss, and $\mathcal{L}^{outer}=({d}^{mem}, {d}^{fid})$ is the two-element vector of memorization and quality objectives respectively  (Sec.~\ref{sec:metrics}). Note that because the outer loop is a multi-objective optimization, $\omega^*$ is not a single model, but the set of mask configurations that constitute the Pareto front dominating others in terms of memorization and quality. Where necessary, we pick a single model from the Pareto front for evaluation by scalarizing the score vector (for example by arithmetic mean) and picking the optimum mask.

\begin{figure*}[ht]
  \centering
  \includegraphics[width=0.7\linewidth]{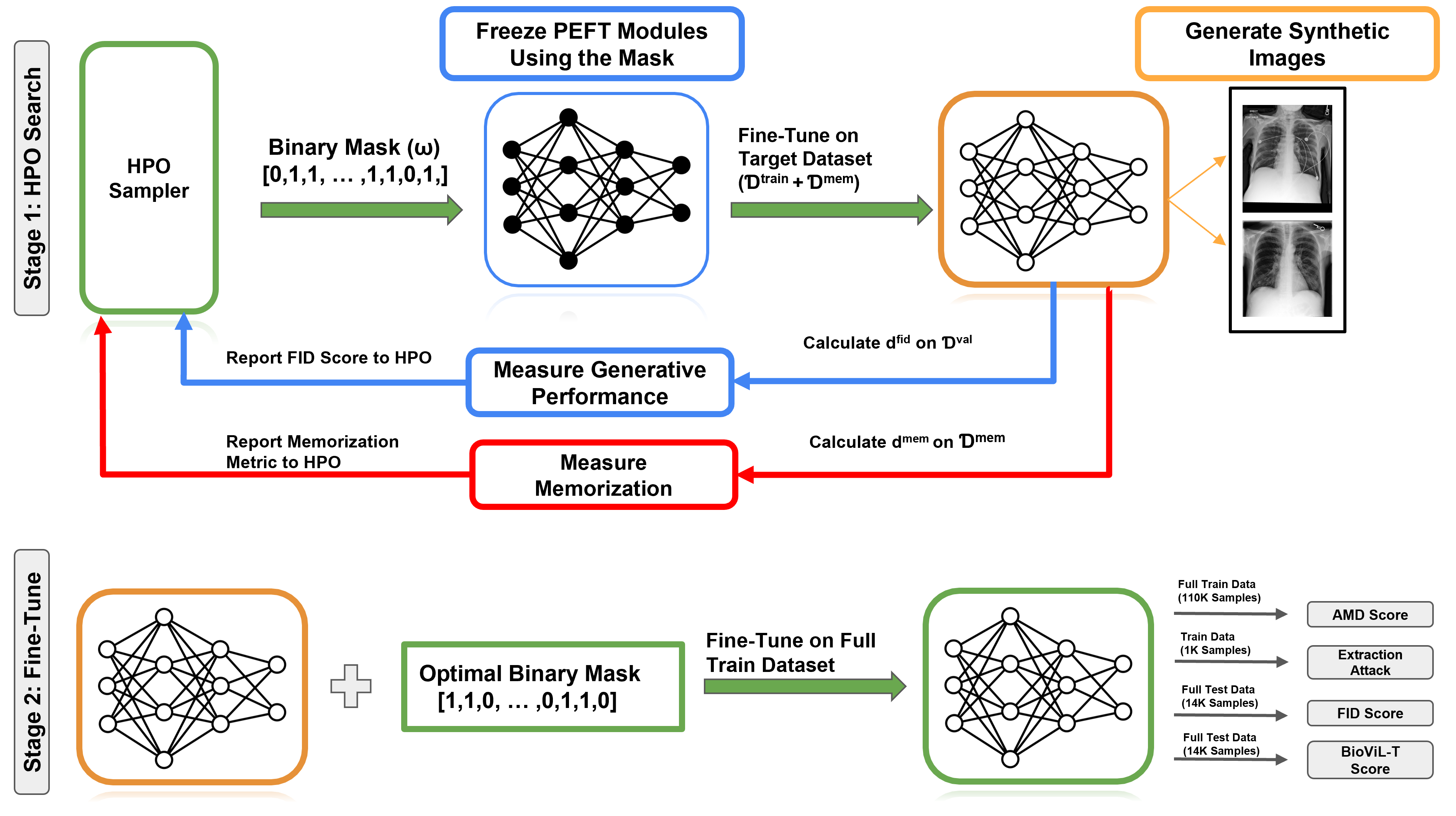}
  \caption{Overall schematic of our framework. \textbf{Stage 1: PEFT Mask Search (top): } We use a subset of the training set to search for the mask that decides which PEFT components of a pre-trained model $\theta$ should be fine-tuned to optimise both generation quality ($d^{fid}$) and memorization ($d^{mem}$). 
  \textbf{Stage 2: Fine-tune with mask (bottom): } The optimal mask from the HPO search is used for fine-tuning on full dataset and final results are reported on the test set ($\mathcal{D}^{test}$).
  }
  \label{fig:schematic}
\end{figure*}

The overall workflow first optimizes the mask using a small proxy dataset (see Sec.~\ref{sub:exp}) for efficient search, and then fixes a specific optimum mask and re-trains on the full training dataset prior to evaluation (Figure~\ref{fig:schematic}).

Additional details on HPO sampling and a detailed analysis of the scalability and convergence of the HPO are presented in the supplementary material (Section 4).
\cut{Outer-loop objective $\mathcal{L}^{outer}$ is optimized via NSGA-II multi-objective algorithm \citep{NGSAII_sampler}, and it uses the crowding distance when evaluating which set of values is better overall.}


\begin{algorithm}[t]
   \caption{Optimizing PEFT for Mitigating Memorization}
   \label{alg:hpo}
\begin{algorithmic}[1]
   \STATE {\bfseries Input:} pre-trained model $\theta_0$, $\alpha$: fine-tuning learning rate, number of trials $T_N$
   \STATE {\bfseries Output:} fine-tuned model $\theta_0+\omega \odot \Delta\phi$ and mask $\omega$
   
   \WHILE{number of completed trials $< T_N$}
   
   \STATE Initialize $\Delta\phi=0$ for $\phi \leftarrow \phi_0 \subset \theta_0$
   \STATE Propose mask $\omega \leftarrow HPO$
  
   \WHILE {not converged}
   \STATE $\Delta\phi \leftarrow \Delta\phi - \alpha\nabla_{\phi} \mathcal{L}^{base}\left(\mathcal{D}^{train}; \theta_0 + \omega \odot \Delta\phi \right)$  $\quad$ // PEFT
   \ENDWHILE
   
   \STATE Evaluate $d^{mem}$ on $\mathcal{D}^{mem}$ and $d^{fid}$ on $\mathcal{D}^{val}$ with model $\theta_0+\omega \odot \Delta\phi$ and report to \textit{HPO} 
   
   \ENDWHILE
   \STATE Return the best masks $\omega$ and fine-tuned models $\theta_0+\omega \odot \Delta\phi$
\end{algorithmic}
\end{algorithm}

\section{Experiments} \label{sec:experiments}

\subsection{Experimental Setup}\label{sub:exp}

\textbf{Architecture:} Our experiments adopted the pre-trained Stable Diffusion (v1.5) \cite{Rombach_2022_CVPR} model implemented in the \textit{Diffusers} package \cite{von-platen-etal-2022-diffusers}. The stable diffusion pipeline consists of three components: (1) a modified U-Net \cite{ronneberger2015u} with self-attention and cross-attention layers for textual guidance, (2) a text-encoder derived from \cite{radford2021learning}, and a variational auto-encoder \cite{kingma2013auto}. For fine-tuning on medical data we follow previous practice \cite{dutt2023parameter, chambon2022roentgen}, and fine-tune only the U-Net while keeping the other components frozen during our experiments. 

\textbf{Baselines: } We compare our approach with different fine-tuning strategies: (1) an off-the-shelf pre-trained Stable Diffusion model that has all of the components frozen, (2) conventional full fine-tuning where every parameter of the U-net is updated, (3) the SV-DIFF PEFT method \cite{han2023svdiff}, (4) the DiffFit PEFT method \cite{xie2023difffit}, and (4) an Attention tuning PEFT method \cite{touvron2022three}. Furthermore, we implement existing state-of-the-art training-time memorization-mitigation methods: (1) Random Word Addition (RWA) \cite{somepalli2023understanding} and threshold mitigation  \cite{wen2024detecting}. Overall our main baselines are a combination of four fine-tuning methods and two mitigation strategies.

\textbf{Search Space and Search Strategy:} Our method is based on optimizing the choice of fine-tuning parameters, which requires defining a search space for $\omega$. {We conduct our experiments on three different parameter search spaces associated with distinct PEFT methods: \textbf{(1)} \textbf{SV-DIFF:} Identifying which attention blocks within the U-Net should include SV-DIFF parameters, \textbf{(2)} \textbf{DiffFit:} Identifying which attention blocks within the U-Net should include DiffFit parameters, \textbf{(3)} \textbf{Attention Tuning:} Determining which attention layers within the U-Net should be fine-tuned. The search space for both SV-DIFF and DiffFit is $\Omega\in\{0,1\}^{13}$, while the search space for attention tuning is $\Omega\in\{0,1\}^{16}$.} We also consider an approxmiate global search over all three search spaces by combining the three pareto fronts from each individual search space. We use the evolutionary search-based NSGA-II Sampler \cite{NGSAII_sampler} as implemented in \textit{Optuna} \cite{akiba2019optuna} to search the space of masks (Eq.~\ref{eq:bilevel-obj}). 

\textbf{Datasets: } We use the MIMIC dataset \cite{johnson2016mimic} that consists of chest X-rays and associated radiology text reports. The MIMIC dataset, because of its scale and rich annotations stands as the standard dataset for performing text-to-image generation in medical image analysis \cite{chambon2022roentgen, dutt2023parameter}.
The dataset was split into train, validation and test splits ensuring no leakage of subjects, resulting in 110K training samples, 14K validation and 14K test samples. Note that that several images in this dataset can be associated with the same text prompt due to similarity in clinical findings.

\textbf{Efficient HPO with a Memorization Subset:} Searching for the optimal parameter subset (mask) across the entire training set is computationally infeasible. Additionally, only a small fraction of images within the training set are actually memorized. Consequently, calculating the memorization metric (Eq.~\ref{eq:searchMetric}, Sec.~\ref{sec:metrics}) would be both noisy and slow if conducted on the entire dataset. To address this, we construct a specific training set, $\mathcal{D}^{train}$, for mask search during the HPO, comprising of \textbf{(1)} 1\% of the original training set and \textbf{(2)} a specialized memorization set, $\mathcal{D}^{mem}$. 

\textbf{Creation of $\mathcal{D}^{mem}$: }We select 100 image-text pairs from the training set and duplicate each 50 times, following the approach in \cite{wen2024detecting,ren2024unveiling}. This duplication forces the model to memorize these specific samples during fine-tuning, providing a clear indication of which pairs are memorized. Unlike natural imaging datasets such as LAION \cite{schuhmann2022laion}, where memorized prompts have been identified in prior studies \cite{webster2023duplication}, no such information exists for medical datasets like MIMIC. Therefore, we explicitly create a subset for this task. We follow a specific heuristic for selecting the 100 the image-text pairs for $\mathcal{D}^{mem}$, detailed in Supplementary material (Section 2).
Following this setup, the HPO training set is small, and we know which images are expected to be memorized. HPO mask search is then driven by the two validation objectives $d^{mem}$ and $d^{fid}$ (Eq.~\ref{eq:bilevel-obj}, Sec.~\ref{sec:metrics}) which are evaluated on the memorization set $\mathcal{D}^{mem}$ and the validation set $\mathcal{D}^{val}$ respectively. 



\subsection{PEFT Mask Choices Explore the Memorization-Generation Tradeoff} \label{sec:capacity_mediates}

\begin{figure*}[ht]
  \centering
  \includegraphics[width=0.7\linewidth]{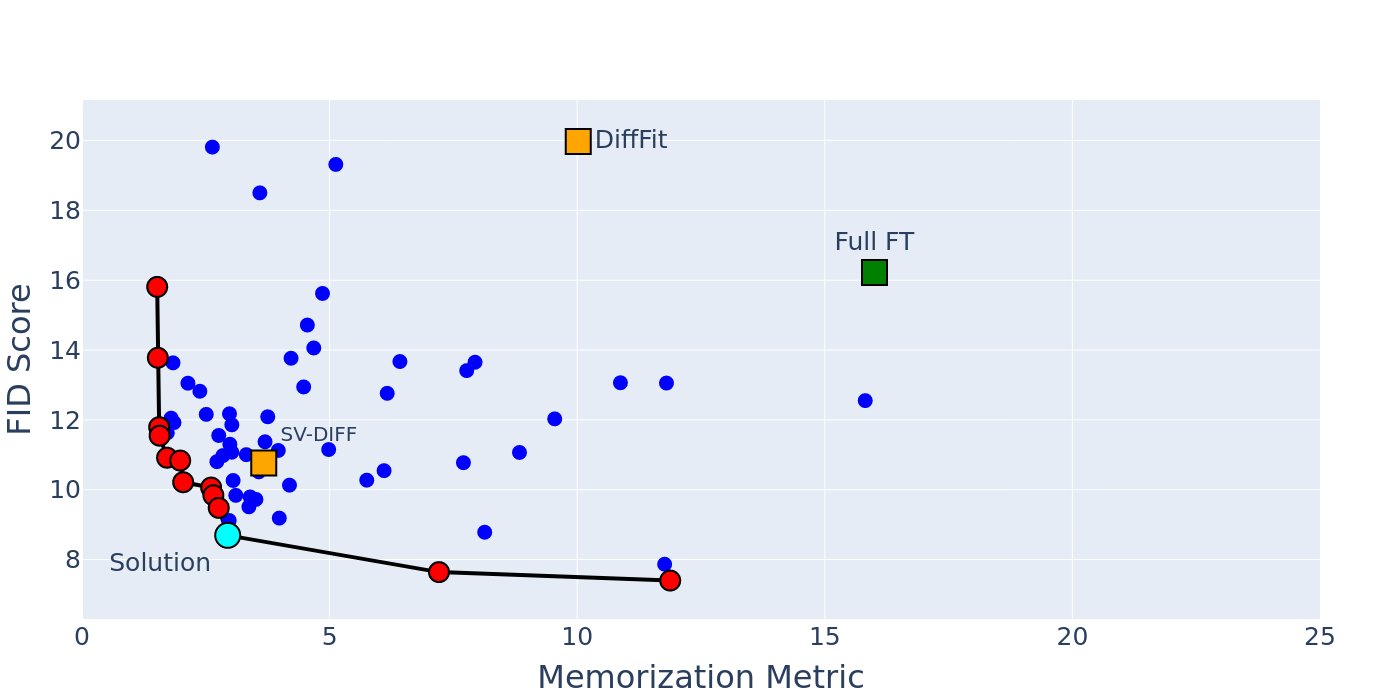}
  \caption{Plot illustrating how model capacity affects the memorization vs. generation quality tradeoff. The HPO search explored various combinations of parameter subsets during fine-tuning (\textcolor{blue}{blue markers}). Each combination results in a different model capacity, generation quality ($d^{fid} \downarrow$) and memorization ($d^{mem}, \downarrow$). The performance of these subsets is compared to two default PEFT configurations (\textcolor{orange}{orange squares}) and full fine-tuning (\textcolor{teal}{green square}). The pareto front of optimal parameter subsets, indicated by the lowest $d^{fid}$ and $d^{mem}$, are marked in \textcolor{red}{red}, while the final optimal solution is marked in \textcolor{cyan}{cyan}.}
  \label{fig:pareto_scatter_plot}
\end{figure*}

We first explore the influence of model capacity on memorization vs. generalization quality tradeoff by recording the FID score ($d^{fid}$) and the memorization metric ($d^{mem}$) for different parameter subsets explored during the HPO search. Each HPO iteration considers a different subset of parameters to freeze or update. The results in Figure~\ref{fig:pareto_scatter_plot} show the combined pareto front from searching within the space of SV-Diff \cite{han2023svdiff}, DiffFit \cite{xie2023difffit} and Attention Tuning \cite{touvron2022three} PEFT options. We also report the performance of the defacto full fine-tuning \cite{dutt2023parameter,kumar2024monitoring}, as well as the vanilla configurations of the PEFT methods.

From the performance scatter plot, we see that certain configurations favour generation quality ($d^{fid} \downarrow$), others might favour reducing memorization ($d^{mem} \downarrow$) while some can perform poorly on both fronts ($d^{fid}$ $\uparrow$, $d^{mem} \uparrow$). However, our framework also discovers the Pareto front of configurations that dominate others ($d^{fid}$ $\downarrow$, $d^{mem} \downarrow$), including configurations that are near optimal for both (bottom left). Importantly, the pareto-front excludes configurations that include the defacto full fine-tuning and PEFT variants.


We select a specific model from the Pareto front for further evaluation by taking the mask configuration with the lowest arithmetic mean of FID score ($d^{fid} \downarrow$) and memorisation score ($d^{mem} \downarrow$). Their scale is relatively comparable and both are minimized, so taking their arithmetic mean is reasonable.

\subsection{Quantitative Comparison vs. State of the Art} \label{sec:HPO_optimization}

\begin{figure}[ht]
  \centering
    \includegraphics[width=0.95\linewidth]{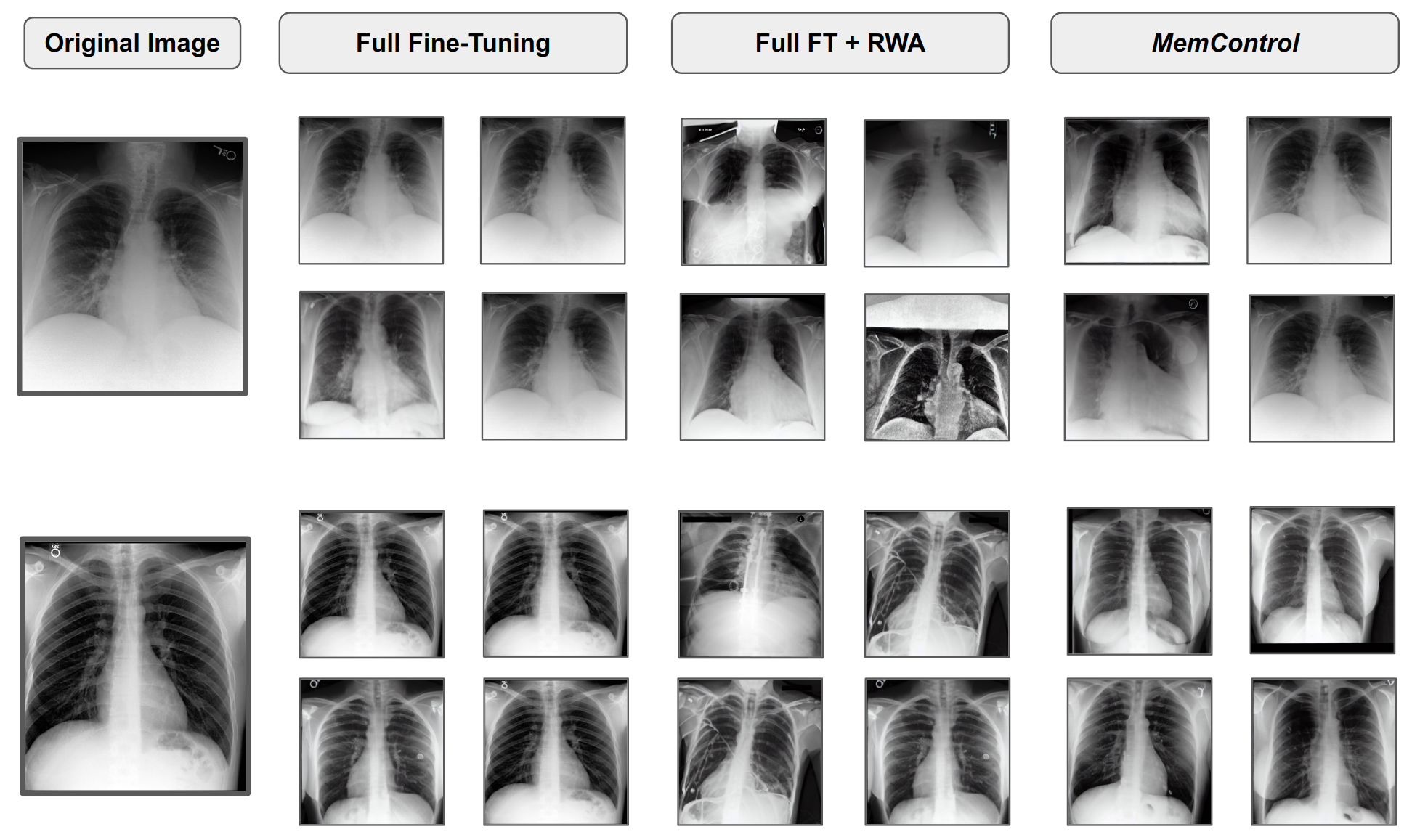}
  \caption{Plot illustrating a qualitative comparison between full fine-tuning (Full FT), Full FT + RWA, and \textit{MemControl}. Full FT (col. 1 \& 2) generates images that are near-replicas of the original training image. Combining Full FT with RWA mitigation strategy (col. 3 \& 4) diversifies the generated images but deteriorates the quality. \textit{MemControl} (col. 5 \& 6) preserves image quality and prevents generating identical images.
  }
  \label{fig:fullft_vs_memcontrol}
\end{figure}

Given the optimal PEFT mask discovered by our framework in Sec.~\ref{sec:capacity_mediates}, we fine-tune using this mask on the full training set, and compare this against competitors. Recall that we consider two types of competitors for memorization mitigation: purpose designed regularisers \cite{somepalli2023understanding,wen2024detecting}, and alternative PEFT methods \cite{han2023svdiff,xie2023difffit}. These mitigation methods are independent and potentially complementary, so we also go beyond prior work in reporting all their combinations as well. Quantitatively, the results in Table \ref{tab:all_results_combined} show that: (1) Despite not being designed to tackle memorization, PEFT methods SV-DIFF \cite{han2023svdiff} and DiffFit \cite{xie2023difffit} outperform state-of-the-art purpose designed mitigation methods such as \cite{wen2024detecting} on several metrics. (2) Our \textit{MemControl} framework achieves the best performance across all metrics. It outperforms both purpose designed mitigation methods, PEFT methods, and their combination across all generation quality and memorization metrics. \textcolor{black}{As a qualitative comparison, we show the synthetic generations in Figure \ref{fig:fullft_vs_memcontrol}. Firstly, it can be directly observed that samples generated with full fine-tuning are near-identical replicas of the ground-truth training images, also highlighted in Figure \ref{fig:memorized_samples}. Additionally, combining full fine-tuning with a mitigation strategy (RWA) negatively impacts the generation performance. Finally, \textit{MemControl} enables high-quality generation and prevents replication simultaneously.}

\begin{table*}[ht]
\centering
\resizebox{0.9\textwidth}{!}{%
\begin{tabular}{@{}lcccc@{}}
\toprule
\textbf{FT Method + Mitigation Strategy}                  & \textbf{FID Score ($\downarrow$)} & \textbf{AMD ($\uparrow$)} & \textbf{BioViL-T Score ($\uparrow$)} & \textbf{Num. Extracted Images ($\downarrow$)} \\ \midrule
Freeze                             & 323.78                            & \textbf{0.801}                        & 0.23             & \textbf{\phantom{00}0}                    \\
Full FT \cite{chambon2022roentgen}                           & \phantom{0}35.79                             & 0.029                       & 0.60             & 356                    \\
Full FT \cite{chambon2022roentgen} + RWA \cite{somepalli2023understanding}                      & \phantom{0}67.48                             & 0.041                       & 0.58       & 320                          \\
Full FT \cite{chambon2022roentgen} + Threshold Mitigation \cite{wen2024detecting}    & \phantom{0}25.43                             & 0.048                        & 0.62       & 298                           \\
SV-DIFF  \cite{han2023svdiff} & \phantom{0}12.77                             & 0.062                        & 0.72       & \phantom{0}63                        \\
SV-DIFF \cite{han2023svdiff} + RWA  \cite{somepalli2023understanding}                    & \phantom{0}41.25                             & 0.059                        & 0.52       & \phantom{0}58                          \\
SV-DIFF \cite{han2023svdiff} + Threshold Mitigation \cite{wen2024detecting}     & \phantom{0}23.64                             & 0.068                       & 0.64       & \phantom{0}45                          \\
DiffFit  \cite{xie2023difffit}                          & \phantom{0}15.19                             & 0.061                        & 0.69       & \phantom{0}69                          \\
DiffFit \cite{xie2023difffit} + RWA  \cite{somepalli2023understanding}                    & \phantom{0}17.18                             & 0.060                        & 0.67       & \phantom{0}64                          \\
DiffFit \cite{xie2023difffit} + Threshold Mitigation \cite{wen2024detecting}    & \phantom{0}15.31                             & 0.067                        & 0.69       & \phantom{0}51                          \\
Attn Tuning  \cite{touvron2022three}        & \phantom{0}40.28                             & 0.031                       & 0.59       & 358                          \\
Attn Tuning \cite{touvron2022three} + RWA   \cite{somepalli2023understanding}               & \phantom{0}49.71                             & 0.035                       & 0.52       & 324                          \\
Attn Tuning \cite{touvron2022three} + Threshold Mitigation \cite{wen2024detecting}  & \phantom{0}25.43                             & 0.041                        & 0.62       & 299                          \\
\rowcolor[HTML]{C0C0C0} 
\textit{\textbf{MemControl (ours)}}                  & \phantom{0}\textbf{11.93}                    & \textbf{0.114}               & \textbf{0.75}   & \phantom{0}\textbf{28}                      \\ \\

\textit{MemControl} + RWA \cite{somepalli2023understanding}                  & \phantom{0}13.10              & 0.121          & \phantom{0}0.69        &     \phantom{0}23        \\
\rowcolor[HTML]{C0C0C0} 
\textbf{\textit{MemControl} + Threshold Mitigation \cite{wen2024detecting}} & \phantom{0}\textbf{11.31}     & \textbf{0.142} & \phantom{0}\textbf{0.75}   &    \phantom{0}\textbf{16}     \\ \bottomrule
\end{tabular}%
}
\caption{Comparative evaluation of generation quality (FID $\downarrow$, BioViL-T Score $\uparrow$) and memorization (AMD $\uparrow$, Extracted Images $\downarrow$) on the MIMIC dataset. Our MemControl outperforms purpose-designed memorization mitigations, standard PEFT methods, and their combinations.\\
Additionally, \textit{MemControl} can be seamlessly integrated with these approaches and effectively complement existing methods to enhance memorization mitigation (bottom two rows). }
\label{tab:all_results_combined}
\end{table*}

\subsection{PEFT Strategy is Transferable} \label{sec:transferability}

One might question if a universal approach to balance quality and memorization across multiple datasets exists. To investigate this, we assess the transferability of our PEFT strategy, initially discovered on the MIMIC medical dataset in Sec.~\ref{sec:capacity_mediates}, to a different dataset, namely Imagenette \cite{Howard_Imagenette_2019}, which serves as a smaller and more manageable subset of ImageNet. Specifically, we fine-tune the Stable Diffusion model on Imagenette using the mask derived from MIMIC.

\begin{table}[h]
\centering
\resizebox{0.48\textwidth}{!}{
\begin{tabular}{@{}lccc@{}}
\toprule
\textbf{FT Method + Mitigation Strategy}                      & \textbf{FID Score ($\downarrow$)} & \textbf{AMD ($\uparrow$)} & \textbf{Num. Extracted Images ($\downarrow$)} \\ \midrule
Freeze                                  & 62.16                                          & \textbf{0.101}                      & \textbf{105}               \\
Full FT                                 & 11.01                                          & 0.044                      & 226               \\
Full FT + RWA                           & \phantom{0}8.72                                           & 0.048           & 198                          \\
Full FT + Threshold Mitigation          & \phantom{0}9.11                                           & 0.052           & 181                          \\
SV-DIFF                                 & \phantom{0}8.10                                           & 0.058           & 172                          \\
SV-DIFF + RWA                           & 13.53                                          & 0.061                & 163                     \\
SV-DIFF + Threshold Mitigation          & \phantom{0}9.38                                          & 0.065            & 158                         \\
DiffFit                                 & \phantom{0}7.40                                            & 0.054          & 144                           \\
DiffFit + RWA                           & 11.94                                          & 0.058                      & 177               \\
DiffFit + Threshold Mitigation          & 12.03                                          & 0.062                      & 135               \\
Attention Tuning                        & \phantom{0}8.04                                           & 0.047           & 194                          \\
Attention Tuning + RWA                  & \phantom{0}9.12                                           & 0.049           &  192                         \\
Attention Tuning + Threshold Mitigation & \phantom{0}9.11                                           & 0.051           & 182                          \\
\rowcolor[HTML]{C0C0C0} 
\textit{\textbf{MemControl (ours)}}     & \phantom{0}\textbf{6.34}                                  & \textbf{0.093}    & \textbf{108} \\ \bottomrule                      
\end{tabular}
}
\caption{Evaluation of generative performance and memorization on Imagenette dataset using parameter subsets learnt from the MIMIC dataset. Our framework leads to the best overall performance in terms of both FID and AMD, showcasing the transferability of strategies from one domain to another.}
\label{tab:imagenette_results}
\end{table}

The results presented in Table \ref{tab:imagenette_results} indicate that our \textit{MemControl} method continues to outperform competitors in both generation quality and memorization. This demonstrates the potential for the transferability of fine-tuning masks across datasets, implying that these masks could be reused for new datasets in the future, eliminating the need for repeated mask searches.

\subsection{Additional Ablations}

\textbf{\textit{MemControl} with existing mitigation mechanisms:} \textcolor{black}{The results in Table \ref{tab:all_results_combined} (bottom two rows)} show that combining \textit{MemControl} with existing mitigation mechanisms leads to improved performance. We do observe a decrease in generation performance in the case of RWA, but its combination with \textit{MemControl} leads to reduced memorization (AMD $ \uparrow$). A detailed analysis of the performance decrease observed with RWA is provided in the supplementary material (Section 2.2). Combining \textit{MemControl} with Threshold Mitigation leads to the best generation and memorization performance overall.


\textbf{\textit{MemControl} can optimize individual PEFT methods:} Our main result in Table~\ref{tab:all_results_combined} applied \textit{MemControl} across all three PEFT search spaces considered. In Table \ref{tab:individual_space} we demonstrate that when applied to the search space associated with a single PEFT method, \textit{MemControl} can substantially improve its performance.

\begin{table}[ht]
\centering
\resizebox{0.48\textwidth}{!}{%
\begin{tabular}{@{}lcccc@{}}
\toprule
\textbf{FT Method + Mitigation Strategy} & \textbf{FID Score} ($\downarrow$)                           & \textbf{AMD}  ($\uparrow$)                                & \textbf{BioViL-T} ($\uparrow$)                           & \textbf{Num. Extracted Images}  ($\downarrow$)             \\ \midrule
SV-DIFF                                  & 12.77                                        & 0.062                                        & 0.72                                         & \phantom{0}63                                           \\
SV-DIFF + RWA                            & 41.25                                        & 0.059                                        & 0.52                                         & \phantom{0}58                                           \\
SV-DIFF + Threshold Mitigation           & 23.64                                        & 0.068                                        & 0.64                                         & \phantom{0}45                                           \\
\rowcolor[HTML]{C0C0C0} 
\textbf{\textit{MemControl} (SV-DIFF Search Space)}        &      \textbf{11.93}                                        &       \textbf{0.114}                                       &       \textbf{0.75}                                       &        \phantom{0}\textbf{28}                                      \\
                                         & \multicolumn{1}{l}{}                         & \multicolumn{1}{l}{}                         & \multicolumn{1}{l}{}                         & \multicolumn{1}{l}{}                         \\
DiffFit                                  & 15.19                                        & 0.061                                        & 0.69                                         & \phantom{0}69                                           \\
DiffFit + RWA                            & 17.18                                        & 0.060                                        & 0.67                                         & \phantom{0}64                                           \\
\rowcolor[HTML]{FFFFFF} 
DiffFit + Threshold Mitigation           & 15.31                                        & 0.067                                        & 0.69                                         & \phantom{0}51                                           \\
\rowcolor[HTML]{C0C0C0} 
\textbf{\textit{MemControl} (DiffFit Search Space)}        & \textbf{12.21} & \textbf{0.086} & \textbf{0.72} & \phantom{0}\textbf{42} \\
                                         & \multicolumn{1}{l}{}                         & \multicolumn{1}{l}{}                         & \multicolumn{1}{l}{}                         & \multicolumn{1}{l}{}                         \\
Attn Tuning                              & 40.28                                        & 0.031                                        & 0.59                                         & 358                                          \\
Attn Tuning + RWA                        & 49.71                                        & 0.035                                        & 0.52                                         & 324                                          \\
Attn Tuning + Threshold Mitigation       & 25.43                                        & 0.041                                        & 0.62                                         & 299                                          \\
\rowcolor[HTML]{C0C0C0} 
\textbf{\textit{MemControl} (Attn Tuning Search Space)}    &   \textbf{22.36}  & \textbf{0.044} & \textbf{0.063} &  \textbf{281}                
\\ \bottomrule
\end{tabular}
}
\caption{Evaluating \textit{MemControl} in individual parameter search spaces. Our method leads to the best performance in the respective search space defined by distinct PEFT methods (SV-DIFF, DiffFit, Attention Tuning) across all evaluation metrics.}
\label{tab:individual_space}
\end{table}
\vspace{-0.2em}

\textbf{Analyzing the Best Solution: } \quad Our best solution relied on a subset of SV-DIFF learnable parameters. We analyze the mask indicating which subset of parameters should be fine-tuned. It indicates necessity of fine-tuning all of the cross-attention layers in the Stable Diffusion U-Net, which strongly aligns with the findings in \cite{han2023svdiff}. Additionally, the mask suggests fine-tuning all of the self-attention layers except for those in down-blocks 2 and 3 and up-blocks 3 and 4. Adopting this strategy corresponds to fine-tuning 0.019\% of the parameters in Stable Diffusion U-Net.

\section{Conclusion} 

In this paper, we conducted the first empirical study of memorization in text-to-image diffusion models applied to medical images. We addressed the challenge of mitigating memorization from the perspective of parameter-efficient fine-tuning with capacity control. Our work demonstrated how the selection of fine-tuning parameter space influences the balance between memorization and generation quality. Through the search for optimal PEFT parameter subsets, we identified configurations that achieve a promising trade-off between generation fidelity and memorization. This introduces a novel approach to addressing memorization in diffusion models, distinct from the current mainstream methods based on regularization and token rewriting.

\section{Acknowledgment}

Raman Dutt is supported by United Kingdom Research and Innovation (grant
EP/S02431X/1), UKRI Centre for Doctoral Training in Biomedical AI at the University of Edinburgh. P. Sanchez thanks additional financial support from the School of Engineering, the University of Edinburgh. S.A. Tsaftaris acknowledges the UK’s Engineering and Physical Sciences Research Council (EPSRC) (grant EP/X017680/1) and
the UKRI AI programme and EPSRC, for CHAI - EPSRC AI Hub for Causality in Healthcare AI with Real Data (grant EP/Y028856/1). 

{\small
\bibliographystyle{ieee_fullname}
\bibliography{PaperForReview}

\begin{thebibliography}{10}\itemsep=-1pt

\bibitem{akbar2023beware}
Muhammad~Usman Akbar, Wuhao Wang, and Anders Eklund.
\newblock Beware of diffusion models for synthesizing medical images-a comparison with gans in terms of memorizing brain mri and chest x-ray images.
\newblock {\em Available at SSRN 4611613}, 2023.

\bibitem{akiba2019optuna}
Takuya Akiba, Shotaro Sano, Toshihiko Yanase, Takeru Ohta, and Masanori Koyama.
\newblock Optuna: A next-generation hyperparameter optimization framework.
\newblock In {\em KDD}, 2019.

\bibitem{bai2021training}
Ching-Yuan Bai, Hsuan-Tien Lin, Colin Raffel, and Wendy Chi-wen Kan.
\newblock On training sample memorization: Lessons from benchmarking generative modeling with a large-scale competition.
\newblock In {\em Proceedings of the 27th ACM SIGKDD conference on knowledge discovery \& data mining}, 2021.

\bibitem{bannur2023learning}
Shruthi Bannur, Stephanie Hyland, Qianchu Liu, Fernando Perez-Garcia, Maximilian Ilse, Daniel~C Castro, Benedikt Boecking, Harshita Sharma, Kenza Bouzid, Anja Thieme, et~al.
\newblock Learning to exploit temporal structure for biomedical vision-language processing.
\newblock In {\em Proceedings of the IEEE/CVF Conference on Computer Vision and Pattern Recognition}, 2023.

\bibitem{basu2023strong}
Samyadeep Basu, Daniela Massiceti, Shell~Xu Hu, and Soheil Feizi.
\newblock Strong baselines for parameter efficient few-shot fine-tuning.
\newblock {\em arXiv}, 2023.

\bibitem{bitfit}
Elad Ben~Zaken, Yoav Goldberg, and Shauli Ravfogel.
\newblock {B}it{F}it: Simple parameter-efficient fine-tuning for transformer-based masked language-models.
\newblock In {\em ACL}, May 2022.

\bibitem{butterick_2023}
Matthew Butterick.
\newblock Stable diffusion litigation · joseph saveri law firm \& matthew butterick, Jan 2023.

\bibitem{carlini2023extracting}
Nicolas Carlini, Jamie Hayes, Milad Nasr, Matthew Jagielski, Vikash Sehwag, Florian Tramer, Borja Balle, Daphne Ippolito, and Eric Wallace.
\newblock Extracting training data from diffusion models.
\newblock In {\em 32nd USENIX Security Symposium (USENIX Security 23)}.

\bibitem{carlini2021extracting}
Nicholas Carlini, Florian Tramer, Eric Wallace, Matthew Jagielski, Ariel Herbert-Voss, Katherine Lee, Adam Roberts, Tom Brown, Dawn Song, Ulfar Erlingsson, et~al.
\newblock Extracting training data from large language models.
\newblock In {\em 30th USENIX Security Symposium (USENIX Security 21)}.

\bibitem{chambon2022roentgen}
Pierre Chambon, Christian Bluethgen, Jean-Benoit Delbrouck, Rogier Van~der Sluijs, Ma{\l}gorzata Po{\l}acin, Juan Manuel~Zambrano Chaves, Tanishq~Mathew Abraham, Shivanshu Purohit, Curtis~P Langlotz, and Akshay Chaudhari.
\newblock Roentgen: vision-language foundation model for chest x-ray generation.
\newblock {\em arXiv:2211.12737}.

\bibitem{chavhan2024memorized}
Ruchika Chavhan, Ondrej Bohdal, Yongshuo Zong, Da Li, and Timothy Hospedales.
\newblock Memorized images in diffusion models share a subspace that can be located and deleted.
\newblock {\em arXiv preprint arXiv:2406.18566}, 2024.

\bibitem{chen2022adaptformer}
Shoufa Chen, Chongjian Ge, Zhan Tong, Jiangliu Wang, Yibing Song, Jue Wang, and Ping Luo.
\newblock Adaptformer: Adapting vision transformers for scalable visual recognition.
\newblock {\em arXiv}, 2022.

\bibitem{torchxrayvision}
Joseph~Paul Cohen, Mohammad Hashir, Rupert Brooks, and Hadrien Bertrand.
\newblock On the limits of cross-domain generalization in automated x-ray prediction.
\newblock In {\em Medical Imaging with Deep Learning}, 2020.

\bibitem{dar2023investigating}
Salman Ul~Hassan Dar, Arman Ghanaat, Jannik Kahmann, Isabelle Ayx, Theano Papavassiliu, Stefan~O Schoenberg, and Sandy Engelhardt.
\newblock Investigating data memorization in 3d latent diffusion models for medical image synthesis.
\newblock In {\em International Conference on Medical Image Computing and Computer-Assisted Intervention}.

\bibitem{NGSAII_sampler}
K. Deb, A. Pratap, S. Agarwal, and T. Meyarivan.
\newblock A fast and elitist multiobjective genetic algorithm: Nsga-ii.
\newblock {\em IEEE Transactions on Evolutionary Computation}.

\bibitem{dhariwal2021diffusion}
Prafulla Dhariwal and Alexander Nichol.
\newblock Diffusion models beat gans on image synthesis.
\newblock {\em Advances in neural information processing systems}.

\bibitem{dutt2024fairtune}
Raman Dutt, Ondrej Bohdal, Sotirios~A. Tsaftaris, and Timothy Hospedales.
\newblock Fairtune: Optimizing parameter efficient fine tuning for fairness in medical image analysis.
\newblock In {\em The Twelfth International Conference on Learning Representations}, 2024.

\bibitem{dutt2023parameter}
Raman Dutt, Linus Ericsson, Pedro Sanchez, Sotirios~A Tsaftaris, and Timothy Hospedales.
\newblock Parameter-efficient fine-tuning for medical image analysis: The missed opportunity.
\newblock {\em arXiv}, 2023.

\bibitem{dutt2022automatic}
Raman Dutt, Dylan Mendonca, Huai~Ming Phen, Samuel Broida, Marzyeh Ghassemi, Judy Gichoya, Imon Banerjee, Tim Yoon, and Hari Trivedi.
\newblock Automatic localization and brand detection of cervical spine hardware on radiographs using weakly supervised machine learning.
\newblock {\em Radiology: Artificial Intelligence}.

\bibitem{fernandez2023privacy}
Virginia Fernandez, Pedro Sanchez, Walter Hugo~Lopez Pinaya, Grzegorz Jacenk{\'o}w, Sotirios~A Tsaftaris, and Jorge Cardoso.
\newblock Privacy distillation: reducing re-identification risk of multimodal diffusion models.
\newblock {\em arXiv:2306.01322}, 2023.

\bibitem{ghalebikesabi2023differentially}
Sahra Ghalebikesabi, Leonard Berrada, Sven Gowal, Ira Ktena, Robert Stanforth, Jamie Hayes, Soham De, Samuel~L Smith, Olivia Wiles, and Borja Balle.
\newblock Differentially private diffusion models generate useful synthetic images.
\newblock {\em arXiv preprint arXiv:2302.13861}, 2023.

\bibitem{han2023svdiff}
Ligong Han, Yinxiao Li, Han Zhang, Peyman Milanfar, Dimitris Metaxas, and Feng Yang.
\newblock Svdiff: Compact parameter space for diffusion fine-tuning.
\newblock {\em arXiv:2303.11305}, 2023.

\bibitem{ho2020denoising}
Jonathan Ho, Ajay Jain, and Pieter Abbeel.
\newblock Denoising diffusion probabilistic models.
\newblock {\em Advances in neural information processing systems}, 2020.

\bibitem{hong2024decoding}
Junyuan Hong, Jinhao Duan, Chenhui Zhang, Zhangheng Li, Chulin Xie, Kelsey Lieberman, James Diffenderfer, Brian Bartoldson, Ajay Jaiswal, Kaidi Xu, et~al.
\newblock Decoding compressed trust: Scrutinizing the trustworthiness of efficient llms under compression.
\newblock {\em arXiv preprint arXiv:2403.15447}, 2024.

\bibitem{Howard_Imagenette_2019}
Jeremy Howard.
\newblock Imagenette: A smaller subset of 10 easily classified classes from imagenet, March 2019.

\bibitem{hu2021lora}
Edward~J Hu, Yelong Shen, Phillip Wallis, Zeyuan Allen-Zhu, Yuanzhi Li, Shean Wang, Lu Wang, and Weizhu Chen.
\newblock Lora: Low-rank adaptation of large language models.
\newblock {\em arXiv:2106.09685}, 2021.

\bibitem{hu2022lora}
Edward~J Hu, yelong shen, Phillip Wallis, Zeyuan Allen-Zhu, Yuanzhi Li, Shean Wang, Lu Wang, and Weizhu Chen.
\newblock Lo{RA}: Low-rank adaptation of large language models.
\newblock In {\em ICLR}, 2022.

\bibitem{huang2017densely}
Gao Huang, Zhuang Liu, Laurens Van Der~Maaten, and Kilian~Q Weinberger.
\newblock Densely connected convolutional networks.
\newblock In {\em Proceedings of the IEEE conference on computer vision and pattern recognition}.

\bibitem{huang2023make}
Rongjie Huang, Jiawei Huang, Dongchao Yang, Yi Ren, Luping Liu, Mingze Li, Zhenhui Ye, Jinglin Liu, Xiang Yin, and Zhou Zhao.
\newblock Make-an-audio: Text-to-audio generation with prompt-enhanced diffusion models.
\newblock In {\em International Conference on Machine Learning}.

\bibitem{jegorova2022survey}
Marija Jegorova, Chaitanya Kaul, Charlie Mayor, Alison~Q O'Neil, Alexander Weir, Roderick Murray-Smith, and Sotirios~A Tsaftaris.
\newblock Survey: Leakage and privacy at inference time.
\newblock {\em IEEE Transactions on Pattern Analysis and Machine Intelligence}.

\bibitem{johnson2016mimic}
Alistair~EW Johnson, Tom~J Pollard, Lu Shen, Li-wei~H Lehman, Mengling Feng, Mohammad Ghassemi, Benjamin Moody, Peter Szolovits, Leo Anthony~Celi, and Roger~G Mark.
\newblock Mimic-iii, a freely accessible critical care database.
\newblock {\em Scientific data}.

\bibitem{karras2019style}
Tero Karras, Samuli Laine, and Timo Aila.
\newblock A style-based generator architecture for generative adversarial networks.
\newblock In {\em Proceedings of the IEEE/CVF conference on computer vision and pattern recognition}, 2019.

\bibitem{kim2022guided}
Heeseung Kim, Sungwon Kim, and Sungroh Yoon.
\newblock Guided-tts: A diffusion model for text-to-speech via classifier guidance.
\newblock In {\em International Conference on Machine Learning}.

\bibitem{kingma2013auto}
Diederik~P Kingma and Max Welling.
\newblock Auto-encoding variational bayes.
\newblock {\em arXiv:1312.6114}, 2013.

\bibitem{kopiczko2024vera}
Dawid~Jan Kopiczko, Tijmen Blankevoort, and Yuki~M Asano.
\newblock Ve{RA}: Vector-based random matrix adaptation.
\newblock In {\em The Twelfth International Conference on Learning Representations}, 2024.

\bibitem{kumar2024monitoring}
Sourav Kumar, Shell~Xu Hu, Timothy Hospedales, Praveer Singh, and Jayashree Kalpathy-cramer.
\newblock Monitoring disease progression with stable diffusion- a visual approach.
\newblock In {\em Medical Imaging with Deep Learning}, 2024.

\bibitem{lian2022scaling}
Dongze Lian, Daquan Zhou, Jiashi Feng, and Xinchao Wang.
\newblock In S. Koyejo, S. Mohamed, A. Agarwal, D. Belgrave, K. Cho, and A. Oh, editors, {\em Advances in Neural Information Processing Systems}.

\bibitem{liu2024dora}
Shih-Yang Liu, Chien-Yi Wang, Hongxu Yin, Pavlo Molchanov, Yu-Chiang~Frank Wang, Kwang-Ting Cheng, and Min-Hung Chen.
\newblock Dora: Weight-decomposed low-rank adaptation.
\newblock {\em arXiv:2402.09353}, 2024.

\bibitem{loshchilov2018decoupled}
Ilya Loshchilov and Frank Hutter.
\newblock Decoupled weight decay regularization.
\newblock In {\em ICLR}, 2018.

\bibitem{maini2023can}
Pratyush Maini, Michael~C Mozer, Hanie Sedghi, Zachary~C Lipton, J~Zico Kolter, and Chiyuan Zhang.
\newblock Can neural network memorization be localized?
\newblock {\em arXiv preprint arXiv:2307.09542}, 2023.

\bibitem{naseh2023memory}
Ali Naseh, Jaechul Roh, and Amir Houmansadr.
\newblock Memory triggers: Unveiling memorization in text-to-image generative models through word-level duplication.
\newblock {\em arXiv:2312.03692}, 2023.

\bibitem{radford2021learning}
Alec Radford, Jong~Wook Kim, Chris Hallacy, Aditya Ramesh, Gabriel Goh, Sandhini Agarwal, Girish Sastry, Amanda Askell, Pamela Mishkin, Jack Clark, et~al.
\newblock Learning transferable visual models from natural language supervision.
\newblock In {\em International conference on machine learning}. PMLR, 2021.

\bibitem{ramesh2022hierarchical}
Aditya Ramesh, Prafulla Dhariwal, Alex Nichol, Casey Chu, and Mark Chen.
\newblock Hierarchical text-conditional image generation with clip latents.
\newblock {\em arXiv:2204.06125}, 2022.

\bibitem{ren2024unveiling}
Jie Ren, Yaxin Li, Shenglai Zen, Han Xu, Lingjuan Lyu, Yue Xing, and Jiliang Tang.
\newblock Unveiling and mitigating memorization in text-to-image diffusion models through cross attention.
\newblock {\em arXiv:2403.11052}, 2024.

\bibitem{rombach2022high}
Robin Rombach, Andreas Blattmann, Dominik Lorenz, Patrick Esser, and Bj{\"o}rn Ommer.
\newblock High-resolution image synthesis with latent diffusion models.
\newblock In {\em Proceedings of the IEEE/CVF conference on computer vision and pattern recognition}.

\bibitem{Rombach_2022_CVPR}
Robin Rombach, Andreas Blattmann, Dominik Lorenz, Patrick Esser, and Bj\"orn Ommer.
\newblock High-resolution image synthesis with latent diffusion models.
\newblock In {\em Proceedings of the IEEE/CVF Conference on Computer Vision and Pattern Recognition (CVPR)}, pages 10684--10695, June 2022.

\bibitem{ronneberger2015u}
Olaf Ronneberger, Philipp Fischer, and Thomas Brox.
\newblock U-net: Convolutional networks for biomedical image segmentation.
\newblock In {\em Medical image computing and computer-assisted intervention--MICCAI 2015: 18th international conference, Munich, Germany, October 5-9, 2015, proceedings, part III 18}. Springer, 2015.

\bibitem{Radioactive_data}
Alexandre Sablayrolles, Matthijs Douze, Cordelia Schmid, and Herve Jegou.
\newblock Radioactive data: tracing through training.
\newblock In {\em Proceedings of the 37th International Conference on Machine Learning}.

\bibitem{schuhmann2022laion}
Christoph Schuhmann, Romain Beaumont, Richard Vencu, Cade Gordon, Ross Wightman, Mehdi Cherti, Theo Coombes, Aarush Katta, Clayton Mullis, Mitchell Wortsman, et~al.
\newblock Laion-5b: An open large-scale dataset for training next generation image-text models.
\newblock {\em Advances in Neural Information Processing Systems}.

\bibitem{sohl2015deep}
Jascha Sohl-Dickstein, Eric Weiss, Niru Maheswaranathan, and Surya Ganguli.
\newblock Deep unsupervised learning using nonequilibrium thermodynamics.
\newblock In {\em International conference on machine learning}.

\bibitem{somepalli2023diffusion}
Gowthami Somepalli, Vasu Singla, Micah Goldblum, Jonas Geiping, and Tom Goldstein.
\newblock Diffusion art or digital forgery? investigating data replication in diffusion models.
\newblock In {\em Proceedings of the IEEE/CVF Conference on Computer Vision and Pattern Recognition}, 2023.

\bibitem{somepalli2023understanding}
Gowthami Somepalli, Vasu Singla, Micah Goldblum, Jonas Geiping, and Tom Goldstein.
\newblock Understanding and mitigating copying in diffusion models.
\newblock {\em Advances in Neural Information Processing Systems}, 36, 2023.

\bibitem{song2021denoising}
Jiaming Song, Chenlin Meng, and Stefano Ermon.
\newblock Denoising diffusion implicit models.
\newblock In {\em International Conference on Learning Representations}, 2021.

\bibitem{song2021scorebased}
Yang Song, Jascha Sohl-Dickstein, Diederik~P Kingma, Abhishek Kumar, Stefano Ermon, and Ben Poole.
\newblock Score-based generative modeling through stochastic differential equations.
\newblock In {\em International Conference on Learning Representations}, 2021.

\bibitem{stoehr2024localizing}
Niklas Stoehr, Mitchell Gordon, Chiyuan Zhang, and Owen Lewis.
\newblock Localizing paragraph memorization in language models.
\newblock {\em arXiv preprint arXiv:2403.19851}, 2024.

\bibitem{touvron2022three}
Hugo Touvron, Matthieu Cord, Alaaeldin El-Nouby, Jakob Verbeek, and Herv{\'e} J{\'e}gou.
\newblock Three things everyone should know about vision transformers.
\newblock In {\em ECCV}, 2022.

\bibitem{trabucco2024effective}
Brandon Trabucco, Kyle Doherty, Max~A Gurinas, and Ruslan Salakhutdinov.
\newblock Effective data augmentation with diffusion models.
\newblock In {\em The Twelfth International Conference on Learning Representations}, 2024.

\bibitem{vignac2023digress}
Clement Vignac, Igor Krawczuk, Antoine Siraudin, Bohan Wang, Volkan Cevher, and Pascal Frossard.
\newblock Digress: Discrete denoising diffusion for graph generation.
\newblock In {\em The Eleventh International Conference on Learning Representations}, 2023.

\bibitem{von-platen-etal-2022-diffusers}
Patrick von Platen, Suraj Patil, Anton Lozhkov, Pedro Cuenca, Nathan Lambert, Kashif Rasul, Mishig Davaadorj, Dhruv Nair, Sayak Paul, William Berman, Yiyi Xu, Steven Liu, and Thomas Wolf.
\newblock Diffusers: State-of-the-art diffusion models.
\newblock \url{https://github.com/huggingface/diffusers}, 2022.

\bibitem{wang2023diagnosis}
Zhenting Wang, Chen Chen, Lingjuan Lyu, Dimitris~N Metaxas, and Shiqing Ma.
\newblock Diagnosis: Detecting unauthorized data usages in text-to-image diffusion models.
\newblock In {\em The Twelfth International Conference on Learning Representations}, 2023.

\bibitem{webster2023duplication}
Ryan Webster, Julien Rabin, Loic Simon, and Frederic Jurie.
\newblock On the de-duplication of laion-2b.
\newblock {\em arXiv:2303.12733}, 2023.

\bibitem{wen2024detecting}
Yuxin Wen, Yuchen Liu, Chen Chen, and Lingjuan Lyu.
\newblock Detecting, explaining, and mitigating memorization in diffusion models.
\newblock In {\em The Twelfth International Conference on Learning Representations}, 2024.

\bibitem{xie2023difffit}
Enze Xie, Lewei Yao, Han Shi, Zhili Liu, Daquan Zhou, Zhaoqiang Liu, Jiawei Li, and Zhenguo Li.
\newblock Difffit: Unlocking transferability of large diffusion models via simple parameter-efficient fine-tuning.
\newblock {\em arXiv}, 2023.

\bibitem{xu2022geodiff}
Minkai Xu, Lantao Yu, Yang Song, Chence Shi, Stefano Ermon, and Jian Tang.
\newblock Geodiff: A geometric diffusion model for molecular conformation generation.
\newblock In {\em International Conference on Learning Representations}, 2022.

\bibitem{yoon2020anonymization}
Jinsung Yoon, Lydia~N Drumright, and Mihaela Van Der~Schaar.
\newblock Anonymization through data synthesis using generative adversarial networks (ads-gan).
\newblock {\em IEEE journal of biomedical and health informatics}, 2020.

\bibitem{zhai2019large}
Xiaohua Zhai, Joan Puigcerver, Alexander Kolesnikov, Pierre Ruyssen, Carlos Riquelme, Mario Lucic, Josip Djolonga, Andre~Susano Pinto, Maxim Neumann, Alexey Dosovitskiy, et~al.
\newblock A large-scale study of representation learning with the visual task adaptation benchmark.
\newblock {\em arXiv:1910.04867}, 2019.

\end{thebibliography}
}


\end{document}